\documentclass[runningheads]{llncs}

\usepackage{eccv}

\usepackage{eccvabbrv}

\usepackage{graphicx}
\usepackage{booktabs}

\usepackage[accsupp]{axessibility}  

\usepackage{hyperref}

\usepackage{orcidlink}

\usepackage{algorithm}
\usepackage{algorithmic}

\def \R {\mathbb{R}}
\def \x {\mathbf{x}}

\def \p {\mathbf{p}}
\def \q {\mathbf{q}}
\def \g {\mathbf{g}}
\def \w {\mathbf{w}}
\def \D {\mathcal{D}}

\newtheorem{prop}{Proposition}

\begin{document}

\title{SeA: Semantic Adversarial Augmentation for Last Layer Features from Unsupervised Representation Learning} 

\titlerunning{Semantic Adversarial Augmentation for Last Layer Features}

\author{Qi Qian\inst{1}\orcidlink{0009-0007-8661-1169}\thanks{Corresponding author} \and
Yuanhong Xu\inst{2}\orcidlink{0009-0006-3238-125x} \and
Juhua Hu\inst{3}\orcidlink{0000-0001-5869-3549}}

\authorrunning{Q.~Qian et al.}

\institute{Alibaba Group, Bellevue, WA 98004, USA \and 
Alibaba Group, Hangzhou, China \and
School of Engineering and Technology, \\University of Washington, Tacoma, WA 98402, USA\\
\email{\{qi.qian, yuanhong.xuyh\}@alibaba-inc.com, juhuah@uw.edu}}

\maketitle

\begin{abstract}
  Deep features extracted from certain layers of a pre-trained deep model show superior performance over the conventional hand-crafted features. Compared with fine-tuning or linear probing that can explore diverse augmentations, \eg, random crop/flipping, in the original input space, the appropriate augmentations for learning with fixed deep features are more challenging and have been less investigated, which degenerates the performance. To unleash the potential of fixed deep features, we propose a novel semantic adversarial augmentation (SeA) in the feature space for optimization. Concretely, the adversarial direction implied by the gradient will be projected to a subspace spanned by other examples to preserve the semantic information. Then, deep features will be perturbed with the semantic direction, and augmented features will be applied to learn the classifier. Experiments are conducted on $11$ benchmark downstream classification tasks with $4$ popular pre-trained models. Our method is $2\%$ better than the deep features without SeA on average. Moreover, compared to the expensive fine-tuning that is expected to give good performance, SeA shows a comparable performance on $6$ out of $11$ tasks, demonstrating the effectiveness of our proposal in addition to its efficiency. Code is available at \url{https://github.com/idstcv/SeA}.
\end{abstract}

\section{Introduction}
\label{sec:intro}
Deep learning can be partially considered as a representation learning method that aims to extract features from raw data directly. By obtaining appropriate representations, a simple linear model can achieve state-of-the-art performance on challenging tasks, \eg, classification~\cite{HeZRS16}, object detection~\cite{RenHG017}, \etc. After the success of deep learning~\cite{KrizhevskySH12}, researchers investigate the representations learned from a large-scale data set, \ie, ImageNet~\cite{RussakovskyDSKS15}. Surprisingly, the deep features extracted from a certain layer of a pre-trained model can outperform hand-crafted features on various downstream tasks~\cite{DonahueJVHZTD14}, demonstrating the efficacy of the data-dependent representation learning mechanism implied by deep learning.

\begin{figure}[t]
\centering
\includegraphics[height = 1.55in]{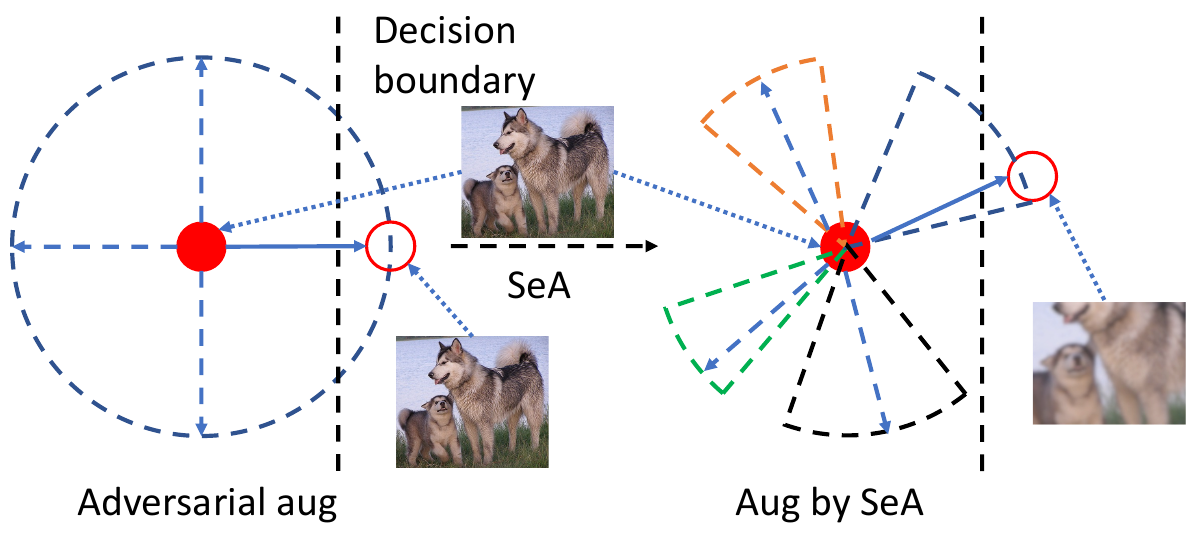}
\caption{Illustration of semantic adversarial augmentation (SeA). The red solid and empty circles denote the original data and its augmentation, respectively. Left: Conventional adversarial augmentation can perturb with arbitrary direction (\eg, augmentation may appear the same as the original); Right: SeA augments examples with semantic directions spanned by features from real data (\eg, different sectors show different subspaces for augmentation, where we can get more semantic meaningful augmentation). }\label{fig:illu}
\end{figure}

Unlike hand-crafted features, representations obtained by deep learning are highly data/task-dependent as illustrated in \cite{Xu000H21}. When optimizing the task from the source domain, the neural network focuses on exploring the patterns related to the specific training task, while ignoring diverse information that has potential for different tasks. Hence, different representations can be learned from different training tasks even with the same training data from the source domain. Conventional methods learn representations with the labels of examples, which capture the knowledge only for the given labels and limit the information in deep features of each example.

To mitigate the problem, fine-tuning parameters of the whole network becomes prevalent for various downstream tasks~\cite{KornblithSL19}. On one hand, fine-tuning can benefit from the prior knowledge in the pre-trained representations. On the other hand, it can further optimize representations with diverse augmentations on the target task. Consequently, fine-tuning works better than learning with fixed representations on downstream tasks.

An important advantage of fine-tuning over learning with fixed deep features is the additional information from semantic augmentations. Obtaining effective augmentations in image space is convenient with semantic operators such as random crop, flipping, \etc. In contrast, obtaining appropriate augmentations in the feature space for deep features becomes challenging, due to the lack of semantic operations. To leverage input space augmentations, a linear classifier can be learned with a frozen backbone by linear probing~\cite{ChenXH21} that generates augmented images for optimization at each iteration.

While fine-tuning and linear probing show promising performance, learning with fixed features is still attractive due to its good properties for real-world applications. First, it only applies deep models to extract features for each original example once, which is applicable for limited computational resources. On the contrary, fine-tuning has the full forward and backward pass and linear probing has the forward pass for each augmented example, which is more expensive for optimization. Second, the same models can be reused for different downstream tasks to extract deep features, while fine-tuning will adjust all parameters in pre-trained models. Thereafter, it has to keep a specific deep model for each task, which becomes intractable for handling hundreds of downstream tasks simultaneously. Finally, given fixed features, learning a linear classifier can be formulated as a convex problem that has the global optimum with a theoretical guarantee~\cite{boyd2004convex}, while preserving the knowledge from the pre-trained model in features. It requires much less tuning efforts than fine-tuning that has to be tuned carefully to avoid the collapse of pre-trained parameters and catastrophic forgetting~\cite{LiuXX00JC022}. Therefore, we focus on fixed deep features extracted after the last pooling layer (\ie, inputs for the last fully-connected layer) in this work. 

Although some existing works~\cite{VermaLBNMLB19,isda} consider augmentation in intermediate layers to help optimize the whole network, little efforts were devoted to the deep features after the last pooling layer. Moreover, the existing feature space augmentation method~\cite{isda} is proposed to complement the input space augmentation. It relies on other augmentation techniques and cannot work well solely.

To tackle the problem, in this work, we propose a novel semantic adversarial augmentation strategy in the feature space of fixed deep features. Concretely, the gradient of each example is computed at first, and then a semantic direction can be observed by projecting the gradient to the subspace spanned by real data. Finally, the features of examples are augmented according to the obtained semantic direction for learning the linear classifier. \cref{fig:illu} illustrates the proposed augmentation method that obtains effective augmentations with only fixed features. Moreover, to further mitigate the information loss in deep features from supervised pre-trained models, those from pre-trained models with different unsupervised pretext tasks are also investigated with an ensemble strategy. Our main contributions can be summarized as follows.
\begin{itemize}
    \item We empirically demonstrate the current performance gap between fine-tuning and learning with fixed deep features. Those deep features extracted from $4$ representative pre-trained models are evaluated on $11$ downstream tasks. 
    \item To improve the performance of deep features, we propose a semantic adversarial augmentation method to obtain appropriate augmentations tailored for fixed features. In addition, a smoothed hinge loss is investigated to demonstrate the augmentation direction explicitly.
    \item Our proposed SeA gains $2$\% accuracy on average over $11$ downstream tasks compared to the baseline using deep features without augmentations. Moreover, compared with fine-tuning that is expected to give good performance, our recipe for learning with deep features can achieve comparable performance on $6$ out of $11$ tasks with way less computational cost, which shows the potential of self-supervised pre-trained deep features and demonstrates both the efficiency and effectiveness of our proposal.
\end{itemize}

\section{Related Work}
\label{sec:related}

\subsection{Deep Features}
Modern deep neural networks consist of multiple layers to exact representations from raw materials. After that, a simple linear model encoded by a fully-connected layer can be attached on top of the representations for classification. After pre-training on a large-scale data set, the obtained neural network can be considered as a feature extractor and a new fully-connected layer can be learned for target tasks. 

When the full label information of data is available, the learning objective is explicit and a supervised representation learning can be conducted by optimizing a conventional classification task, which is equivalent to optimizing the triplet loss as in distance metric learning~\cite{QianSSHTLJ19}. The obtained deep features can outperform hand-crafted features on different applications, \eg, classification~\cite{DonahueJVHZTD14}, distance metric learning~\cite{QianJZL15}, \etc. Besides, some work aimed to obtain robust representations for different downstream tasks within the framework of classification~\cite{QianHL20}. 

Without any label information, the learning objective varies in unsupervised learning. First, each instance can be considered as an individual class and the representations can be learned by instance discrimination~\cite{ChenK0H20,He0WXG20}. In addition, clustering can be applied to capture the relationship between different instances. A coarse-grained classification task defined on clusters can be leveraged to optimize representations~\cite{CaronMMGBJ20,coke}. Finally, other pretext tasks beyond classification also demonstrate the effective representation after pre-training~\cite{GrillSATRBDPGAP20}. Compared with supervised representation learning, the different objectives in unsupervised learning can learn various semantic information even from the same data set. In this work, we will systematically study representations from different pre-trained models and illustrate the performance gap compared to fine-tuning the whole neural network.

\subsection{Augmentation in Feature Space}
Due to the over-parameterization property of deep neural networks, augmentation is essential for training deep models to avoid over-fitting~\cite{KrizhevskySH12}. Given an image, a perturbed copy can be observed by standard image operations such as random crop, flipping, color jitter, \etc~\cite{HeZRS16}. The augmented images have a large variance while preserving the semantic information of the original image, \ie, an image of a cat is still a cat after augmentation, which helps train effective models with the additional information.

However, augmentation in feature space becomes more challenging due to the lack of semantic preserving operators. Some methods consider the augmentation of features from intermediate layers to help train the whole deep neural networks in different applications~\cite{VolpiMSM18,ChenCGWWLW22,VermaLBNMLB19,isda}. For example, \cite{VolpiMSM18} aims to learn feature augmentation with a feature generator for unsupervised domain adaptation. \cite{ChenCGWWLW22} tries to augment intermediate feature maps with adversarial feature moments in batch normalization~\cite{IoffeS15} for efficient training. Manifold mixup~\cite{VermaLBNMLB19} proposes to apply the original mixup~\cite{ZhangCDL18} to features from multiple layers for augmentation, which also includes the layer investigated in this work. However, their study shows that the gain of mixup is mainly from the original image space and feature space of early stages, while that from the feature space of the last layer is negligible, which is consistent with our observation in the ablation study. ISDA~\cite{isda} considers the semantic augmentation in feature space but it has to obtain the semantic direction with input space augmentations, which is complementary to input space augmentations but cannot work as the sole augmentation for deep features well. On the contrary, we propose SeA to project the adversarial direction with features of real data points. Moreover, our proposal is tailored for fixed features and is different from existing works that are for optimizing the whole network.

\section{Semantic Adversarial Augmentation}
\label{sec:method}

We start the analysis with a standard classification framework. Let $\{x_i, y_i\}_{i=1}^n$ denote the training data set and $\{f_k\}_{k=1}^K$ contains a set of $K$ pre-trained models, which can be pre-trained with different learning objectives on different data sets. In this work, we directly use the representations extracted from pre-trained deep models and learn a simple linear model on top of it. Representations from different deep models are concatenated as the final representation for each example. Given $x_i$, the representation will be extracted as 
\begin{eqnarray}\label{eq:cat}
\x_i = [f_1(x_i),\dots,f_K(x_i)]\in\R^d
\end{eqnarray}
where ${d=\sum_k d_k}$ and $d_k$ is the dimension of the representation from the $k$-th pre-trained deep model.

With the above fixed deep features, a classification model can be learned by minimizing the empirical risk with the appropriate regularization as
\begin{eqnarray}\label{eq:obj}
\min_{\w} \sum_i\ell(\x_i, y_i; \w) + \frac{\tau}{2}\|\w\|_F^2
\end{eqnarray}
where $\tau$ is the weight for $L_2$ regularization and $\w$ denotes parameters of the classification model, which is a linear classifier with the prediction probability of the $i$-th example on the $j$-th class as
\begin{eqnarray}
p_{i,j} = \frac{\exp(\x_i^\top \w_j)}{\sum_{c=1}^C \exp(\x_i^\top \w_c)}
\end{eqnarray}
$C$ is the number of classes and $\ell$ is a loss function for learning that will be discussed later.

\subsection{Semantic Direction}

In a standard fine-tuning pipeline, an image can be augmented by multiple random perturbations. When only deep features are available, it is hard to adopt existing techniques to generate appropriate augmented examples, since the semantic direction in the feature space is hard to capture. Manifold mixup~\cite{VermaLBNMLB19} shows that randomly augmenting features from the last output layers has little improvement compared to the augmentation in input image space. To facilitate the performance of deep features, we investigate the semantic adversarial direction for augmentation.

First, for the $i$-th example and its representation $\x_i$, we consider its adversarial direction that can be obtained by maximizing the loss function~\cite{GoodfellowSS14}
\begin{eqnarray}
\max_{\x:\|\x-\x_i\|_2\leq \gamma} \ell(\x_i, y_i; \w)
\end{eqnarray}
Note that the gradient indicates the ideal direction for the adversarial perturbation, and a standard adversarial example can be obtained by gradient ascent as
\begin{eqnarray}\label{eq:adv}
\hat{\x}_i = \x_i + \eta \nabla_{\x_i}\ell
\end{eqnarray}
However, the gradient direction may not be semantically informative. \cite{GoodfellowSS14} shows that the loss can be substantially increased with the adversarial example generated from the gradient, while the appearance is almost the same as the original image as illustrated in \cref{fig:illu}. Therefore, the conventional adversarial learning method helps improve the robustness to the adversarial attack but may be inappropriate for regularizing the generic learning task that requires additional information from diverse images.

To obtain the semantic direction, we propose to project the gradient direction to the subspace spanned by real data points. Concretely, let $\{\x_j\}_{j=1}^b$ denote a mini-batch of data and $\g_i = \nabla_{\x_i}\ell$ indicate the gradient of $\x_i$. For the $i$-th image, a semantic adversarial direction $\hat{\g}_i$ consists of representations from data as 
\begin{eqnarray}\label{eq:seg}
\hat{\g}_i = \sum_{j:j\neq i} q_j \x_j;\quad \q^* = \arg\min_{\q\in\Delta} \D(\hat{\g}_i, \g_i)
\end{eqnarray}
where $\D(\cdot,\cdot)$ is a distance function and $\Delta$ is a simplex as $\Delta = \{\q\in\R^{b-1}|\sum_j q_j=1;\forall j, q_j\geq 0\}$. With the projection, we aim to find an adversarial direction in a subspace consisting of original data points as illustrated in \cref{fig:illu}.

The last challenge is to obtain $\q$ efficiently with a distance function. By adopting the squared Euclidean distance, which is a standard distance measurement, the optimization problem for $\q$ can be written as
\begin{eqnarray}
\min_{\q\in\Delta} \|\sum_{j:j\neq i} q_j \x_j - \g_i\|_2^2 - \alpha H(\q)
\end{eqnarray}
where $H(\q)$ is the entropy of $\q$ that helps improve the robustness to different batches.
When normalizing features such that $\|\x_j\|_2 = \|\g_i\|_2=1$ to obtain the direction without the influence from the magnitude, the problem can be upper-bounded as
\begin{prop}
With unit length variables $\{\x_j\}$ and $\g_i$, we have
\begin{eqnarray}
\|\sum_{j:j\neq i} q_j \x_j - \g_i\|_2^2 \leq 2 - 2\sum_j q_j\x_j^\top \g_i
\end{eqnarray}
\end{prop}
The detailed proof can be found in the appendix.

By rearranging the terms, we can maximize the lower-bound of the original problem as
\begin{eqnarray}\label{eq:q}
\max_{\q\in\Delta} \sum_{j:j\neq i} q_j\x_j^\top \g_i + \alpha H(\q)
\end{eqnarray}

According to the K.K.T. condition~\cite{boyd2004convex}, $\q$ has a closed-form solution.
\begin{prop}
The problem in Eqn.~\ref{eq:q} has the optimal solution as
\begin{eqnarray}
q_j = \frac{\exp(\x_j^\top \g_i/\alpha)}{Z};\quad Z = \sum_{k:k\neq i} \exp(\x_k^\top \g_i/\alpha)
\end{eqnarray}
\end{prop}

Given the semantic adversarial direction, the target example with semantic perturbation can be obtained as 
\begin{eqnarray}\label{eq:aug}
\tilde{\x}_i = \Pi(\x_i + \eta \Pi(\hat{\g}_i)) 
\end{eqnarray}
where $\Pi(\cdot)$ normalizes the vector to the unit length if required and $\eta$ denotes the step size for augmentation. Compared with the adversarial perturbation in Eqn.~\ref{eq:adv}, the augmentation in Eqn.~\ref{eq:aug} projects the gradient to the direction consisting of real data points, which can capture the semantic information in the feature space effectively. Alg.~\ref{alg:1} summarizes the proposed method.

\begin{algorithm}[t]
\caption{\textbf{Se}mantic \textbf{A}dversarial Augmentation (SeA) for Given Features}
\begin{algorithmic}[1]
\STATE {\bf Input:} Dataset $\{x_i,y_i\}$, pre-trained models $\{f_k\}$, iterations $T$, linear model $\w$, $\alpha$, $\tau$, $\eta$
\STATE Extract deep features by $\{f_k\}$ for all examples as $\{\x_i\}$
\FOR{$t = 1,\cdots,T$}
\STATE Receive a mini-batch of examples $\{\x_i,y_i\}_{i=1}^b$
\STATE Obtain augmented examples $\{\tilde{\x}_i,y_i\}$ as in Eqn.~\ref{eq:aug}
\STATE Optimize $\w$ by SGD: $\w = \w - \eta_w( \frac{1}{b}\sum_{i}^b\nabla_\w\ell(\tilde{\x}_i,y_i; \w)+\tau \w)$ 
\ENDFOR
\RETURN $\w$
\end{algorithmic}\label{alg:1}
\end{algorithm}

\subsection{Illustration of Adversarial Direction}
In this subsection, we will generalize the cross entropy loss to help illustrate the proposed augmentation strategy.

The standard multi-class hinge loss~\cite{CrammerS01} for convex optimization can be written as
\begin{eqnarray}
\ell(\x_i, y_i; \w) = \max\{0, \delta+\max_{c\neq y_i}\x_i^\top \w_c - \x_i^\top \w_{y_i}\}
\end{eqnarray}
where $\delta$ is a pre-defined margin.

Unlike convex optimization, cross-entropy loss is prevalent in deep learning, which is a smooth function and can help accelerate the convergence~\cite{boyd2004convex}. Therefore, we propose to obtain a smoothed hinge loss by introducing a distribution over logits from different classes $\p\in\Delta$ where $\Delta$ is the simplex.

First, the original hinge loss is equivalent to 
\begin{eqnarray}
\ell(\x_i, y_i; \w)=\max_{\p\in \Delta} p_{y_i} \x_i^\top \w_{y_i}+\sum_{c\neq y_i} p_c(\delta+\x_i^\top \w_c) - \x_i^\top \w_{y_i}
\end{eqnarray}

According to the analysis for cross entropy loss~\cite{QianSSHTLJ19}, the loss can be smoothed by adding an entropy regularization for $\p$ as
\begin{align}\label{eq:p}
&\ell(\x_i, y_i; \w)=\max_{\p\in \Delta} p_{y_i} \x_i^\top \w_{y_i}+\sum_{c\neq y_i} p_c(\delta+\x_i^\top \w_c)  +\lambda H(\p) - \x_i^\top \w_{y_i}
\end{align}
where $H(\p)$ denotes the entropy of $\p$ and $\lambda$ is the coefficient.

\begin{prop}
The loss function in Eqn.~\ref{eq:p} is equivalent to
\begin{align}\label{eq:loss}
\ell(\x_i, y_i; \w) = -\lambda \log \frac{\exp(\x_i^\top \w_{y_i}/\lambda)}{\exp(\x_i^\top \w_{y_i}/\lambda) +\sum_{c\neq y_i}\exp((\x_i^\top \w_c+\delta)/\lambda)}
\end{align}
\end{prop}

\textbf{Remark} It is obvious that the popular cross entropy loss is a special case of Eqn.~\ref{eq:loss} by letting $\delta=0$ and $\lambda=1$. Our analysis connects the hinge loss in conventional methods to the popular loss function in deep learning.

Finally, since deep features are fixed, we can illustrate the gradient direction with the proposed smoothed loss function explicitly.
\begin{prop}\label{prop:direction}
Given the loss function in Eqn.~\ref{eq:loss}, the gradient of $\x_i$ is
\begin{eqnarray}
\nabla_{\x_i}\ell(\x_i, y_i; \w) = \sum_{c=1}^C p_c \w_c - \w_{y_i}
\end{eqnarray}
where $p_c = \left\{\begin{array}{cc}\frac{\exp((\x_i^\top \w_c+\delta)/\lambda)}{Z}&c\neq y_i\\\frac{\exp(\x_i^\top \w_c/\lambda)}{Z}&c=y_i\end{array}\right.$ and $Z = \exp(\x_i^\top \w_{y_i}/\lambda)+\sum_{c\neq y_i}\exp((\x_i^\top \w_c+\delta)/\lambda)$.
\end{prop}
\textbf{Remark} The gradient direction in the feature space indicates that an adversarial perturbation should be close to the primary directions of other classes, while far away from the direction of the corresponding class.

\section{Experiments}
\label{sec:exp}

To demonstrate our proposed method, we adopt state-of-the-art and widely used pre-trained models to extract deep features for evaluation. It should be noted that only two main architectures, including ResNet-50~\cite{HeZRS16} and vision transformer (ViT)~\cite{DosovitskiyB0WZ21}, are prevalently applied in self-supervised learning~\cite{ChenXH21, GrillSATRBDPGAP20,coke,HeCXLDG22}. Since ViT shows a worse classification performance than ResNet-50 with the frozen backbone as demonstrated in~\cite{HeCXLDG22}, we will focus on ResNet-50 with different public pre-trained parameters in the experiment. Specifically, one supervised pre-trained ResNet-50 and three self-supervised pre-trained ResNet-50 are applied for feature extraction. All of these models are pre-trained on ImageNet-1K~\cite{RussakovskyDSKS15} with different learning objectives. The details of different models can be found in the appendix. The accuracy of the supervised model and that of linear probing for unsupervised models are summarized in \cref{ta:model}. 

\begin{table}[!ht]
\centering
\caption{Performance of $4$ pre-trained models on ImageNet.}\label{ta:model}
\begin{tabular}{|l|c|l|c|}\hline
Model&\#Pre-training Epochs&Objective&Acc\%\\\hline
Supervised~\cite{HeZRS16} (S)& 90& $1,000$-class classification&76.6\\\hline
MoCo-v3~\cite{ChenXH21} (M)& 1,000&instance discrimination&74.6\\\hline
BYOL~\cite{GrillSATRBDPGAP20} (B)& 1,000&regression&74.3\\\hline
CoKe~\cite{coke} (C)& 1,000&cluster discrimination&74.9\\\hline
\end{tabular}
\end{table}

$11$ diverse downstream data sets are applied for evaluation, including Aircraft~\cite{maji2013fine}, Caltech101~\cite{fei2004learning}, Stanford Cars~\cite{krause20133d}, CIFAR-10~\cite{krizhevsky2009learning}, CIFAR-100~\cite{krizhevsky2009learning}, CUB200-2011 (Birds)~\cite{wah2011caltech}, Describable Textures Dataset (DTD)~\cite{cimpoi2014describing}, Flowers~\cite{NilsbackZ08}, Food101~\cite{BossardGG14}, Oxford-IIIT Pet (Pets)~\cite{parkhi2012cats}, and Sun397~\cite{XiaoHEOT10}. We follow the evaluation protocol in \cite{GrillSATRBDPGAP20}. Concretely, all models search hyper-parameters on the provided/generated validation set in each downstream task, and the standard metric on the provided test set is reported. Besides, mean per-class accuracy is reported on Aircraft, Caltech101, Flowers, and Pets, while Top-1 accuracy is utilized for other data sets. More details can be found in \cite{GrillSATRBDPGAP20}.

Each model is fine-tuned with SGD by $100$ epochs for sufficient training. The batch size is $256$ and the momentum is $0.9$. The learning rate is searched in a range of $7$ logarithmically-spaced values between $10^{-4}$ and $10^{-1}$. Weight decay is optional, for which if it is applied, the value will be searched with the same setting between $10^{-6}$ and $10^{-3}$. The standard augmentation, \ie, random crop, random horizontal flipping, is applied as in most existing fine-tuning pipelines.

For a fair comparison, we also apply SGD to learn the linear classifier with fixed deep features using the same batch size and momentum. Unlike fine-tuning, we have the constant learning rate for our method, which is searched in $\{2^{i}\}_{i=-2}^{i=3}$, while weight decay is searched in $\{0, 10^{-6}, 10^{-5},10^{-4}\}$. For other parameters, we have $\alpha$, $\lambda$, $\delta$ and $\eta$ searched in $\{0.01,0.02,0.05\}$, $\{0.05,0.1,1\}$, $\{0, 1\}$, and $\{0,0.2,0.4,0.8\}$, respectively. 

For each image, we extract the output of the final pooling layer as deep features. Features from an individual model are normalized to the unit length. If multiple models are exploited, features from different models are concatenated as in Eqn.~\ref{eq:cat} and the combined feature vector is further normalized to the unit length. The only augmentation for training with fixed features is the proposed semantic adversarial augmentation. To approximate the direction of the gradient for each example, a mini-batch of data is leveraged to optimize the problem in Eqn.~\ref{eq:q}. The projection for normalization is also applied to the augmented examples.

\subsection{Ablation Study}
Before the experiments on downstream tasks, we first investigate the effect of parameters in the proposed augmentation. The ablation study is conducted on CIFAR-100 with features concatenated from $4$ pre-trained models and the accuracy is reported for comparison.

\subsubsection{Effect of Step Size for Augmentation}
\begin{figure}[!ht]
\centering
\begin{minipage}{0.31\textwidth}
\centering
\includegraphics[height = 1.1in]{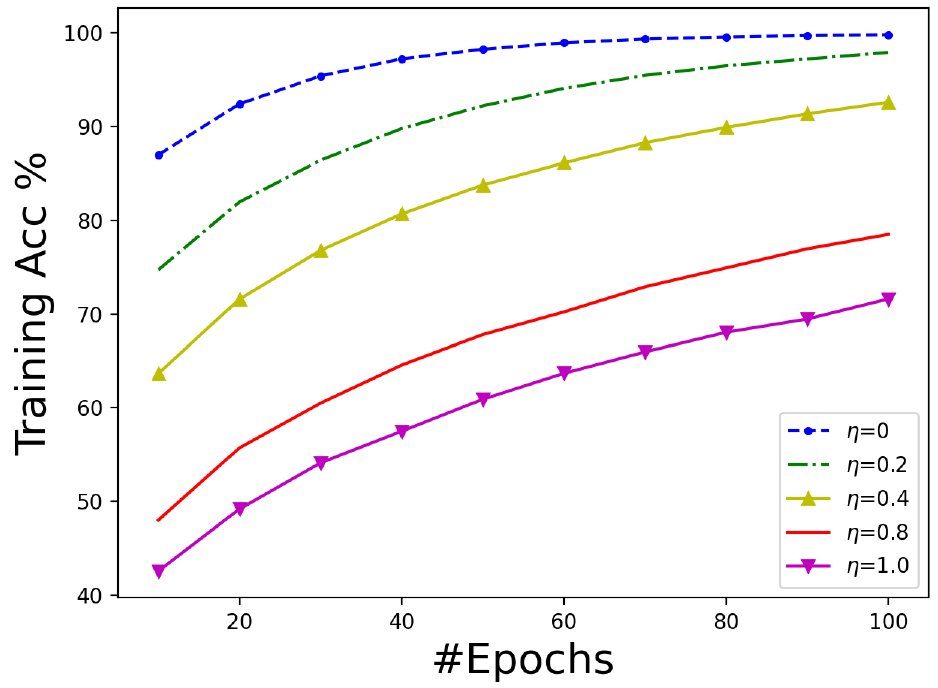}
\caption{Training accuracy of different $\eta$.}\label{fig:eta_train}
\end{minipage}
\begin{minipage}{0.31\textwidth}
\centering
\includegraphics[height = 1.1in]{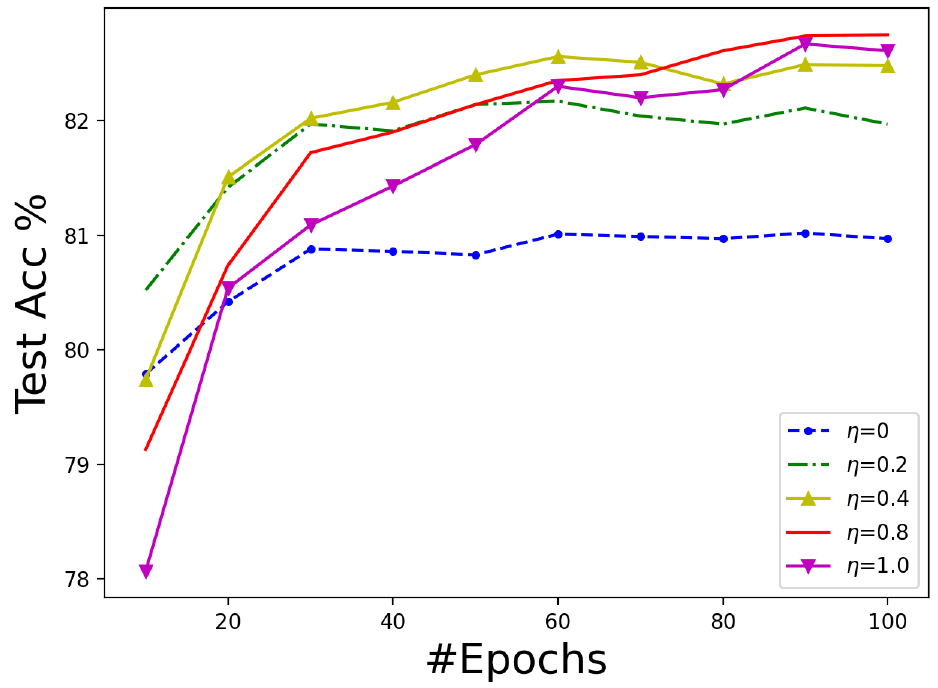}
\caption{Test accuracy of different $\eta$.}\label{fig:eta_test}
\end{minipage}
\begin{minipage}{0.35\textwidth}
\centering
\includegraphics[height = 1.1in]{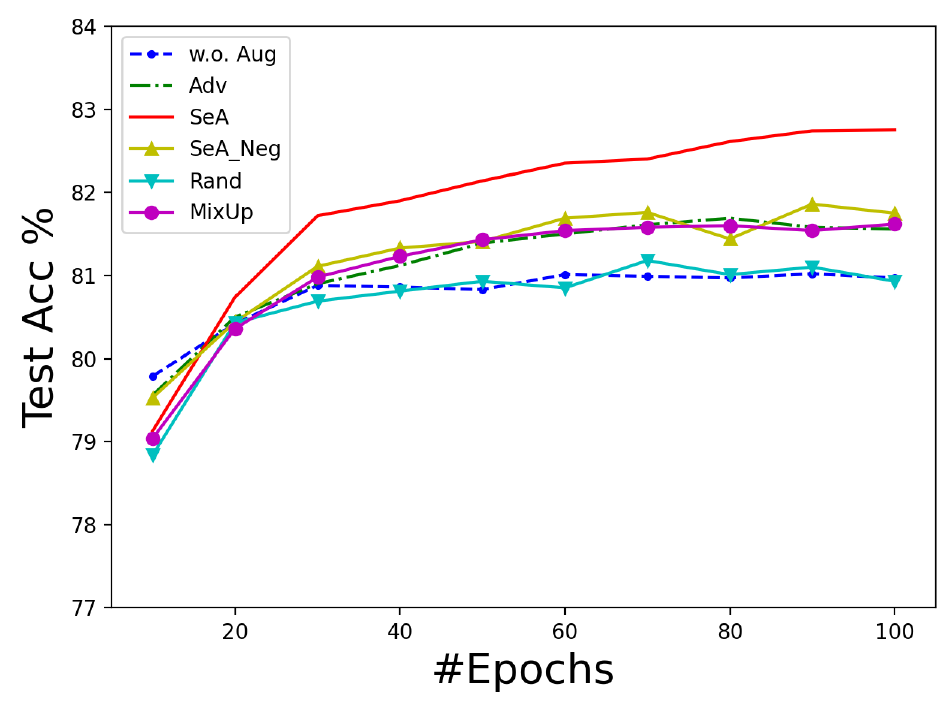}
\caption{Test accuracy of various augmentation directions.}\label{fig:direction}
\end{minipage}
\end{figure}

First, we evaluate the step size $\eta$ in the semantic adversarial augmentation. For SeA, we fix $\lambda=0.1$, $\delta=0$, $\alpha=0.01$, and the learning rate as $1$, according to the performance on the validation set. The weight of the augmentation $\eta$ varies in $\{0,0.2,0.4,0.6,0.8,1.0\}$ and the performance is summarized in \cref{fig:eta_train}-\cref{fig:eta_test}.

When $\eta=0$, there is no augmentation applied for training. Hence, we can observe that the training accuracy exceeds $92\%$ after only $20$ epochs. Due to the over-fitting on the training set, the test accuracy of this baseline is only $81.1\%$ on CIFAR-100. By gradually increasing the step size for the augmentation, the training accuracy is decreased as expected, while the test accuracy increases. It demonstrates that the proposed semantic adversarial augmentation strategy can effectively mitigate the over-fitting problem for learning with fixed deep features. By setting an appropriate step size, the test accuracy can be improved to $82.9\%$, which helps shrink the gap to fine-tuning.

\subsubsection{Effect of Directions for Augmentation}
Different directions can be adopted for generating augmentation. Besides the proposed semantic adversarial augmentations (SeA), four variants are included in the comparison.
\begin{itemize}
    \item Adv: the original gradient direction without projection as in Eqn.~\ref{eq:adv}.
    \item SeA\_Neg: similar to the gradient direction in Proposition~\ref{prop:direction} but getting rid of the direction from the target class and keeping the direction of $\sum_c^C p_c\w_c$.
    \item Rand: a uniformly random direction within subspace spanned by a mini-batch of data, \ie, generating $\q$ in Eqn.~\ref{eq:seg} as a random vector.
    \item Manifold mixup: a random semantic direction indicated by an example~\cite{VermaLBNMLB19}.
\end{itemize}

The step size for Adv is very sensitive and is searched in $\{0.005,0.01,0.02,0.05\}$, while the step size for others is searched in $\{0.2,0.4,0.8,1.0\}$.

\cref{fig:direction} shows the comparison on CIFAR-100. First, we can observe that most augmentation directions except the random one can improve the performance over the baseline without augmentation. It confirms that augmentation is important for better generalization even for learning from fixed representations. Then, the method with the random augmentation direction performs similarly to the baseline, which shows that it is challenging to find an effective direction for augmentation in feature space. However, with the original gradient direction directly as in Adv, the training loss will be significantly increased even with a small step size and only a step size that is much smaller than SeA can obtain an applicable model. This is because that the gradient direction only aims to maximize the loss while ignoring the data distribution in the feature space. By projecting the gradient to a data-dependent subspace, SeA and SeA\_Neg can achieve a better performance than the baseline without a sophisticated setup for the step size. Moreover, by removing the direction from the target class, the accuracy is degenerated by about $1\%$. The comparison demonstrates that keeping the completed gradient direction is essential for obtaining appropriate augmentation. Finally, a random semantic direction indicated by examples as in mixup is worse than SeA with a clear margin.

\subsubsection{Comparison with Linear Probing}
Besides learning with fixed features, linear probing is another way to obtain the linear classifier with a frozen backbone. Compared with our method, it can obtain appropriate augmentations in image space. However, it has to forward each augmentation for learning and the cost is much higher than us.

To illustrate the effectiveness of the proposed augmentation method, we compare the performance between linear probing and our proposal in \cref{ta:lb}. For a fair comparison, all methods optimize features from a single model of MoCo. It shows that the proposed augmentation can outperform the augmentation in the input image space for learning the target classifier.

\begin{table}[!ht]
\centering
\begin{minipage}[t]{0.44\textwidth}
\centering
\caption{Comparison of accuracy (\%) on CIFAR-100. All methods utilize a single model from MoCo.}\label{ta:lb}
\begin{tabular}{|c|c|c|}\hline
baseline&linear probing& SeA\\\hline
77.6&77.8&79.0\\\hline
\end{tabular}
\end{minipage}
\begin{minipage}[t]{0.5\textwidth}
\centering
\caption{Comparison of running time. SeA utilize deep features from $4$ models while other methods optimize only a single model.}\label{ta:time}
\begin{tabular}{|l|c|c|c|c|c|}\hline
Methods&fine-tuning&linear probing& SeA\\\hline
Time (s)&14,383&4,437&40\\\hline
\end{tabular}
\end{minipage}
\end{table}

\subsubsection{Running Time} Finally, we compare the running time of fine-tuning, linear probing, and SeA using fixed deep features in \cref{ta:time}. Note that with features from $4$ models, the dimension of the feature vector is $8,192$. However, optimizing the combined features is still much more efficient than fine-tuning a single model or linear probing with $2,048$ final features. All methods train the corresponding model by $100$ epochs and the running time is measured on a single V100 GPU. Fine-tuning costs $14,383$ (sec) for the entire training. In contrast, SeA only costs $40$ (sec) to obtain the optimal linear model. The efficiency of SeA using deep features implies its applicability for limited-resource scenarios. With efficient optimization, we can repeat experiments of SeA by multiple times. The standard deviation is only $0.03\%$ on CIFAR-100 over 5 runs, which shows the stability of convex optimization. Although linear probing eliminates the cost of the backward pass, that of the forward pass at each iteration is still expensive. Considering that linear probing has a similar running cost as fine-tuning but with worse performance, it will not be included in the comparison over other downstream tasks. More ablation experiments can be found in the appendix.

\subsection{Comparison on Downstream Tasks}
Now we evaluate features from different pre-trained models and the ensemble to demonstrate the effectiveness of SeA on downstream tasks. The comparison is summarized in \cref{ta:tasks}. The linear model with fixed features from different models is denoted by the initial of the corresponding pre-trained model, \ie, S, M, B, and C stand for Supervised, MoCo, BYOL, and CoKe, respectively.

\begin{table}[!ht]
\centering
\caption{Comparison of accuracy ($\%$) on downstream tasks. The best performance within each group is underlined while the global one is in bold.}\label{ta:tasks}
\resizebox{\linewidth}{!}{
\begin{tabular}{|l|c|c|c|c|c|c|c|c|c|c|c|}\hline
Methods&Aircraft&Birds&Caltech&Cars&CIFAR10&CIFAR100&DTD&Flowers&Food&Pets&SUN\\\hline
\multicolumn{12}{|l|}{\textit{Fine-tuning:}}\\\hline
Supervised&87.3&\underline{\textbf{82.6}}&94.6&91.9&97.8&87.4&74.7&\underline{\textbf{97.7}}&88.4&\underline{93.6}&63.6\\
MoCo-v3&87.5&81.5&\underline{94.8}&91.7&\underline{\textbf{98.3}}&87.4&76.3&96.4&88.2&92.2&\underline{65.4}\\
BYOL&88.2&82.0&94.0&91.4&98.0&87.5&76.8&96.8&\underline{\textbf{88.5}}&91.7&63.4\\
CoKe&\underline{\textbf{89.0}}&80.6&94.4&\underline{\textbf{92.1}}&\underline{\textbf{98.3}}&\underline{\textbf{88.3}}&\underline{78.6}&96.7&\underline{\textbf{88.5}}&91.9&64.7\\\hline
\multicolumn{12}{|l|}{\textit{Fixed deep features with SeA:}}\\\hline
S&37.4&	66.5&	92.6&	45.8&	89.8&	71.3&	72.6&	88.1&	67.8&	92.3&	59.2\\
M&61.5&	71.4&	95.4&	69.7&	93.4&	79.0&	76.4&	94.6&	74.9&	90.7&	64.2\\
B&60.0&	66.4&	94.2&	64.2&	91.9&	76.5&	75.8&	95.3&	74.0&	89.7&	61.9\\
C&57.8&	67.0&	94.3&	63.4&	93.0&	78.5&	77.8&	95.6&	76.1&	91.0&	65.3\\\hline
S+M&62.0&	76.3&	96.1&	70.4&	94.4&	80.4&	77.1&	96.3&	77.8&	93.1&	66.3\\
S+B&60.7&	73.7&	95.7&	67.5&	93.4&	79.1&	77.9&	96.3&	77.4&	93.4&	65.5\\
S+C&58.6&	73.7&	95.6&	66.0&	93.5&	80.3&	79.1&	96.2&	78.6&	93.6&	66.9\\
M+B&65.7&	72.5&	95.7&	72.7&	94.6&	80.8&	78.2&	96.2&	77.6&	91.9&	65.5\\
M+C&65.2&	72.2&	95.7&	73.1&	94.6&	81.4&	77.9&	96.5&	78.3&	92.2&	66.6\\
B+C&64.0&	70.1&	95.2&	70.1&	93.9&	80.7&	78.8&	96.4&	78.3&	91.1&	66.4\\\hline
S+M+B&66.2&	76.5&	96.4&	74.2&	94.9&	81.8&	78.6&	96.9&	79.5&	93.3&	67.0\\
S+M+C&65.8&	76.7&	96.3&	73.1&	95.1&	81.9&	78.8&	96.9&	79.9&	93.6&	67.7\\
S+B+C&63.7&	75.4&	96.2&	71.4&	94.4&	82.1&	79.2&	97.0&	79.8&	\underline{\textbf{93.8}}&	67.7\\
M+B+C&67.2&	73.2&	95.9&	74.7&	\underline{95.2}&	82.3&	78.9&	96.8&	79.4&	92.0&	67.2\\\hline
S+M+B+C&\underline{67.4}&	\underline{76.8}&	\underline{\textbf{96.3}}&	\underline{75.4}&	\underline{95.2}&	\underline{82.9}&	\underline{\textbf{79.8}}&	\underline{97.2}&	\underline{80.8}&	\underline{\textbf{93.8}}& \underline{\textbf{68.2}}\\\hline
\end{tabular}}
\end{table}

First, the performance of fine-tuning supervised and unsupervised models varies in different domains. For domains closely related to ImageNet (\eg, Birds, Flowers, Pets), the supervised pre-trained model shows better performance than the unsupervised ones, while unsupervised models surpass the supervised counterpart when the domain gap between pre-training and fine-tuning is large. The phenomenon indicates that representations from supervised/unsupervised models capture different patterns and suggests that combining the information from multiple models may handle downstream tasks better. 

Second, when having the fixed representation from only a single model with SeA, the performance is already close to fine-tuning on certain data sets, \eg, Caltech, DTD, and SUN. However, the information learned from a single model is biased toward the specific pre-training task. Without fine-tuning, the fixed representation only preserves limited patterns from the source domain, which is insufficient for different target domains. 

To mitigate the limitation of individual pre-training tasks, we consider combining the representations from multiple models, which can be done by a simple feature concatenation operator. Compared with features from a single model, the accuracy of all downstream tasks is improved. When having fixed representations from all of $4$ models, a linear classifier learned with SeA can outperform standard fine-tuning on $4$ tasks (\ie, Caltech, DTD, Pets, and SUN) and achieve comparable performance on $2$ tasks (\ie, CIFAR-10 and Flowers). It illustrates that collecting representations from multiple pre-training tasks can obtain complementary patterns that help mitigate the gap to different downstream tasks. 

Moreover, it can be observed that ``S+C'' performs better than ``S+M'' on DTD, while it is worse on Birds. The ensemble of all models, \ie, ``S+M+B+C'', shows the best performance on most tasks. Since the models in evaluation are pre-trained with different learning objectives, we find that the learned deep features complement each other. Therefore, more can be better. Note that ensemble can also be applied to fine-tuning. However, the computational cost will significantly increase depending on the number of models, which becomes unaffordable for real applications. In addition, its ensemble can still be worse than deep features. For example, an ensemble of fine-tuning on DTD achieves $78.9\%$ that is about $1\%$ worse than SeA.

Although the result of SeA is promising, fixed features perform noticeably worse than fine-tuning on two tasks, \eg, Aircraft and Cars. It is because that the gap between the source domain and the target domain is too large. For example, ImageNet contains limited generic images for cars, while the target task is more fine-grained and consists of cars with different make, models, and years. In this case, a larger model pre-trained on a larger data set covering more diverse domains (\eg, JFT-3B~\cite{Zhai0HB22}) may help improve the performance of SeA. 

\subsubsection{Comparison on Ensemble of Features}
Then, we illustrate the effectiveness of SeA by comparing it to the baseline without SeA and existing feature-space augmentation methods in \cref{ta:gain}, which includes the semantic augmentation method, \ie, ISDA~\cite{isda} and a robust adversarial augmentation method, \ie, ME-ADA~\cite{ada}. The experiment is conducted with the ensembled features from $4$ models for all methods. The hyper-parameters in ISDA and ME-ADA are searched for the desired result.

\begin{table}[!ht]
\centering
\caption{Comparison of SeA and benchmark methods on S+M+B+C. ``Gain'' denotes the improvement over the Baseline.}\label{ta:gain}
\resizebox{\linewidth}{!}{
\begin{tabular}{|l|c|c|c|c|c|c|c|c|c|c|c|c|}\hline
Methods&Aircraft&Birds&Caltech&Cars&CIFAR10&CIFAR100&DTD&Flowers&Food&Pets&SUN&Avg\\\hline
Baseline&63.6&	73.4&	95.4&	69.2&	94.8&	81.1&	79.4&	96.2&	79.5&	93.1&	66.9&81.1\\
ISDA&63.3&	73.6&	95.9&	69.4&	94.9&	81.3&	79.6&	95.6&	79.6&	93.2&	67.0&81.2\\
ME-ADA&64.4&	74.9&	95.7&	70.9&	95.0&	81.6&	79.8&	95.8&	80.3&	93.2&	67.4&81.7\\
SeA&67.4&	76.8&	96.3&	75.4&	95.2&	82.9&	79.8&	97.2&	80.8&	93.8& 68.2&83.1\\\hline
Gain&3.8&	3.4&	0.9&	6.2&	0.4&	1.8&	0.4&	1.0&	1.3&	0.7&	1.3&2.0\\\hline
\end{tabular}}
\end{table}

First, compared with the baseline, SeA improves the performance of deep features over all data sets, and the average gain is up to $2\%$ as denoted by ``Gain''. The consistent improvement shows that augmentation is important for learning with fixed deep features and the proposed augmentation helps mitigate the over-fitting problem effectively. Second, while both SeA and ISDA focus on semantic augmentation, ISDA relies on input space augmentations to obtain the semantic directions, while our method leverages features from other examples. Therefore, when input space augmentations are unavailable for deep features, ISDA cannot beat the baseline and it shows that our method is more appropriate for learning without input space augmentations. In addition, ME-ADA improves the vanilla adversarial augmentation by mitigating its sensitivity to the gradient direction. However, the optimization only considers the features from the target example. By incorporating the information from other examples in the same batch, the proposed method can outperform ME-ADA with a margin of $1.4\%$ on average.

\section{Conclusion}
\label{sec:conclude}
With the development of unsupervised representation learning, features with diverse information can be obtained by learning with different pretext tasks. Hence, we aim to investigate state-of-the-art performance of pre-trained deep features in this work. By introducing the novel semantic adversarial augmentation, we can show that learning with fixed features can achieve comparable performance to fine-tuning the whole network, but with way less cost. Investigating the efficacy of SeA in pre-training and fine-tuning can be our future work.

\section*{Limitations}
The proposed augmentation algorithm is for deep features from pre-trained models. Therefore, it requires access to existing pre-trained models. While many models are publicly available, their license may limit the application.

%
%
\bibliographystyle{splncs04}
\bibliography{sea}

\begin{thebibliography}{10}
\providecommand{\url}[1]{\texttt{#1}}
\providecommand{\urlprefix}{URL }
\providecommand{\doi}[1]{https://doi.org/#1}

\bibitem{BossardGG14}
Bossard, L., Guillaumin, M., Gool, L.V.: Food-101 - mining discriminative
  components with random forests. In: Fleet, D.J., Pajdla, T., Schiele, B.,
  Tuytelaars, T. (eds.) ECCV. Lecture Notes in Computer Science, vol.~8694, pp.
  446--461. Springer (2014)

\bibitem{boyd2004convex}
Boyd, S., Boyd, S.P., Vandenberghe, L.: Convex optimization. Cambridge
  university press (2004)

\bibitem{CaronMMGBJ20}
Caron, M., Misra, I., Mairal, J., Goyal, P., Bojanowski, P., Joulin, A.:
  Unsupervised learning of visual features by contrasting cluster assignments.
  In: Larochelle, H., Ranzato, M., Hadsell, R., Balcan, M., Lin, H. (eds.)
  NeurIPS (2020)

\bibitem{ChenCGWWLW22}
Chen, T., Cheng, Y., Gan, Z., Wang, J., Wang, L., Liu, J., Wang, Z.:
  Adversarial feature augmentation and normalization for visual recognition.
  Trans. Mach. Learn. Res.  \textbf{2022} (2022)

\bibitem{ChenK0H20}
Chen, T., Kornblith, S., Norouzi, M., Hinton, G.E.: A simple framework for
  contrastive learning of visual representations. In: ICML. Proceedings of
  Machine Learning Research, vol.~119, pp. 1597--1607. {PMLR} (2020)

\bibitem{ChenXH21}
Chen, X., Xie, S., He, K.: An empirical study of training self-supervised
  vision transformers. In: ICCV. pp. 9620--9629. {IEEE} (2021)

\bibitem{cimpoi2014describing}
Cimpoi, M., Maji, S., Kokkinos, I., Mohamed, S., Vedaldi, A.: Describing
  textures in the wild. In: CVPR. pp. 3606--3613 (2014)

\bibitem{CrammerS01}
Crammer, K., Singer, Y.: On the algorithmic implementation of multiclass
  kernel-based vector machines. J. Mach. Learn. Res.  \textbf{2},  265--292
  (2001)

\bibitem{DonahueJVHZTD14}
Donahue, J., Jia, Y., Vinyals, O., Hoffman, J., Zhang, N., Tzeng, E., Darrell,
  T.: Decaf: {A} deep convolutional activation feature for generic visual
  recognition. In: ICML. {JMLR} Workshop and Conference Proceedings, vol.~32,
  pp. 647--655. JMLR.org (2014)

\bibitem{DosovitskiyB0WZ21}
Dosovitskiy, A., Beyer, L., Kolesnikov, A., Weissenborn, D., Zhai, X.,
  Unterthiner, T., Dehghani, M., Minderer, M., Heigold, G., Gelly, S.,
  Uszkoreit, J., Houlsby, N.: An image is worth 16x16 words: Transformers for
  image recognition at scale. In: ICLR. OpenReview.net (2021)

\bibitem{fei2004learning}
Fei-Fei, L., Fergus, R., Perona, P.: Learning generative visual models from few
  training examples: An incremental bayesian approach tested on 101 object
  categories. In: CVPR workshop. pp. 178--178. IEEE (2004)

\bibitem{GoodfellowSS14}
Goodfellow, I.J., Shlens, J., Szegedy, C.: Explaining and harnessing
  adversarial examples. In: Bengio, Y., LeCun, Y. (eds.) ICLR (2015)

\bibitem{GrillSATRBDPGAP20}
Grill, J., Strub, F., Altch{\'{e}}, F., Tallec, C., Richemond, P.H.,
  Buchatskaya, E., Doersch, C., Pires, B.{\'{A}}., Guo, Z., Azar, M.G., Piot,
  B., Kavukcuoglu, K., Munos, R., Valko, M.: Bootstrap your own latent - {A}
  new approach to self-supervised learning. In: Larochelle, H., Ranzato, M.,
  Hadsell, R., Balcan, M., Lin, H. (eds.) NeurIPS (2020)

\bibitem{HalkoMT11}
Halko, N., Martinsson, P., Tropp, J.A.: Finding structure with randomness:
  Probabilistic algorithms for constructing approximate matrix decompositions.
  {SIAM} Rev.  \textbf{53}(2),  217--288 (2011)

\bibitem{HeCXLDG22}
He, K., Chen, X., Xie, S., Li, Y., Doll{\'{a}}r, P., Girshick, R.B.: Masked
  autoencoders are scalable vision learners. In: CVPR. pp. 15979--15988. {IEEE}
  (2022)

\bibitem{He0WXG20}
He, K., Fan, H., Wu, Y., Xie, S., Girshick, R.B.: Momentum contrast for
  unsupervised visual representation learning. In: CVPR. pp. 9726--9735.
  Computer Vision Foundation / {IEEE} (2020)

\bibitem{HeZRS16}
He, K., Zhang, X., Ren, S., Sun, J.: Deep residual learning for image
  recognition. In: CVPR. pp. 770--778. {IEEE} Computer Society (2016)

\bibitem{IoffeS15}
Ioffe, S., Szegedy, C.: Batch normalization: Accelerating deep network training
  by reducing internal covariate shift. In: Bach, F.R., Blei, D.M. (eds.) ICML.
  {JMLR} Workshop and Conference Proceedings, vol.~37, pp. 448--456. JMLR.org
  (2015)

\bibitem{KornblithSL19}
Kornblith, S., Shlens, J., Le, Q.V.: Do better imagenet models transfer better?
  In: CVPR. pp. 2661--2671. Computer Vision Foundation / {IEEE} (2019)

\bibitem{krause20133d}
Krause, J., Stark, M., Deng, J., Fei-Fei, L.: 3d object representations for
  fine-grained categorization. In: ICCV workshop. pp. 554--561 (2013)

\bibitem{krizhevsky2009learning}
Krizhevsky, A., Hinton, G., et~al.: Learning multiple layers of features from
  tiny images  (2009)

\bibitem{KrizhevskySH12}
Krizhevsky, A., Sutskever, I., Hinton, G.E.: Imagenet classification with deep
  convolutional neural networks. In: Bartlett, P.L., Pereira, F.C.N., Burges,
  C.J.C., Bottou, L., Weinberger, K.Q. (eds.) NeurIPS. pp. 1106--1114 (2012)

\bibitem{LiuXX00JC022}
Liu, Z., Xu, Y., Xu, Y., Qian, Q., Li, H., Ji, X., Chan, A.B., Jin, R.:
  Improved fine-tuning by better leveraging pre-training data. In: NeurIPS
  (2022)

\bibitem{maji2013fine}
Maji, S., Rahtu, E., Kannala, J., Blaschko, M., Vedaldi, A.: Fine-grained
  visual classification of aircraft. arXiv preprint arXiv:1306.5151  (2013)

\bibitem{NilsbackZ08}
Nilsback, M., Zisserman, A.: Automated flower classification over a large
  number of classes. In: ICVGIP. pp. 722--729. {IEEE} Computer Society (2008)

\bibitem{parkhi2012cats}
Parkhi, O.M., Vedaldi, A., Zisserman, A., Jawahar, C.: Cats and dogs. In: CVPR.
  pp. 3498--3505. IEEE (2012)

\bibitem{QianHL20}
Qian, Q., Hu, J., Li, H.: Hierarchically robust representation learning. In:
  CVPR. pp. 7334--7342. Computer Vision Foundation / {IEEE} (2020)

\bibitem{QianJZL15}
Qian, Q., Jin, R., Zhu, S., Lin, Y.: Fine-grained visual categorization via
  multi-stage metric learning. In: CVPR. pp. 3716--3724. {IEEE} Computer
  Society (2015)

\bibitem{QianSSHTLJ19}
Qian, Q., Shang, L., Sun, B., Hu, J., Li, H., Jin, R.: Softtriple loss: Deep
  metric learning without triplet sampling. In: ICCV. pp. 6449--6457. {IEEE}
  (2019)

\bibitem{coke}
Qian, Q., Xu, Y., Hu, J., Li, H., Jin, R.: Unsupervised visual representation
  learning by online constrained k-means. In: CVPR. pp. 16619--16628. {IEEE}
  (2022)

\bibitem{RenHG017}
Ren, S., He, K., Girshick, R.B., Sun, J.: Faster {R-CNN:} towards real-time
  object detection with region proposal networks. {IEEE} Trans. Pattern Anal.
  Mach. Intell.  \textbf{39}(6),  1137--1149 (2017)

\bibitem{RussakovskyDSKS15}
Russakovsky, O., Deng, J., Su, H., Krause, J., Satheesh, S., Ma, S., Huang, Z.,
  Karpathy, A., Khosla, A., Bernstein, M.S., Berg, A.C., Fei{-}Fei, L.:
  Imagenet large scale visual recognition challenge. Int. J. Comput. Vis.
  \textbf{115}(3),  211--252 (2015)

\bibitem{VermaLBNMLB19}
Verma, V., Lamb, A., Beckham, C., Najafi, A., Mitliagkas, I., Lopez{-}Paz, D.,
  Bengio, Y.: Manifold mixup: Better representations by interpolating hidden
  states. In: Chaudhuri, K., Salakhutdinov, R. (eds.) ICML. Proceedings of
  Machine Learning Research, vol.~97, pp. 6438--6447. {PMLR} (2019)

\bibitem{VolpiMSM18}
Volpi, R., Morerio, P., Savarese, S., Murino, V.: Adversarial feature
  augmentation for unsupervised domain adaptation. In: CVPR. pp. 5495--5504.
  Computer Vision Foundation / {IEEE} Computer Society (2018)

\bibitem{wah2011caltech}
Wah, C., Branson, S., Welinder, P., Perona, P., Belongie, S.: The caltech-ucsd
  birds-200-2011 dataset  (2011)

\bibitem{isda}
Wang, Y., Huang, G., Song, S., Pan, X., Xia, Y., Wu, C.: Regularizing deep
  networks with semantic data augmentation. {IEEE} Trans. Pattern Anal. Mach.
  Intell.  \textbf{44}(7),  3733--3748 (2022)

\bibitem{XiaoHEOT10}
Xiao, J., Hays, J., Ehinger, K.A., Oliva, A., Torralba, A.: {SUN} database:
  Large-scale scene recognition from abbey to zoo. In: CVPR. pp. 3485--3492.
  {IEEE} Computer Society (2010)

\bibitem{Xu000H21}
Xu, Y., Qian, Q., Li, H., Jin, R., Hu, J.: Weakly supervised representation
  learning with coarse labels. In: ICCV. pp. 10573--10581. {IEEE} (2021)

\bibitem{Zhai0HB22}
Zhai, X., Kolesnikov, A., Houlsby, N., Beyer, L.: Scaling vision transformers.
  In: CVPR. pp. 1204--1213. {IEEE} (2022)

\bibitem{ZhangCDL18}
Zhang, H., Ciss{\'{e}}, M., Dauphin, Y.N., Lopez{-}Paz, D.: mixup: Beyond
  empirical risk minimization. In: ICLR. OpenReview.net (2018)

\bibitem{ada}
Zhao, L., Liu, T., Peng, X., Metaxas, D.N.: Maximum-entropy adversarial data
  augmentation for improved generalization and robustness. In: Larochelle, H.,
  Ranzato, M., Hadsell, R., Balcan, M., Lin, H. (eds.) NeurIPS (2020)

\end{thebibliography}
\appendix
\section{Theoretical Analysis}
\subsection{Proof of Proposition~1}
\begin{proof}
With Cauchy-Schwarz inequality, we have
\[\|\sum_{j:j\neq i} q_j \x_j \|_2\leq \sum_j q_j\|\x_j\|_2=\sum_j q_j=1\]
\end{proof}

\subsection{Proof of Proposition~3}
\begin{proof}
The equivalent form is directly from the closed-form solution of $\p$ as
\begin{eqnarray}\label{eq:pc}
p_c = \left\{\begin{array}{cc}\frac{\exp((\x_i^\top \w_c+\delta)/\lambda)}{Z}&c\neq y_i\\\frac{\exp(\x_i^\top \w_c/\lambda)}{Z}&c=y_i\end{array}\right.
\end{eqnarray}
where $Z = \exp(\x_i^\top \w_{y_i}/\lambda)+\sum_{c\neq y_i}\exp((\x_i^\top \w_c+\delta)/\lambda)$.
\end{proof}

\section{Experiments}
\subsection{Statistics of Pre-trained Models}

To obtain the supervised one, we follow the standard training pipeline on ImageNet~\cite{HeZRS16} with cosine learning rate decay that shows better performance than the stage-wise decay in the public model\footnote{\url{https://pytorch.org/vision/stable/models.html}}. MoCo-v3\footnote{\url{https://github.com/facebookresearch/moco-v3}}~\cite{ChenXH21}, BYOL\footnote{\url{https://github.com/deepmind/deepmind-research/tree/master/byol}}~\cite{GrillSATRBDPGAP20} and CoKe\footnote{\url{https://github.com/idstcv/CoKe}}~\cite{coke} are adopted as representative self-supervised pre-trained models with the objective of instance discrimination, regression, and cluster discrimination. The publicly available models are applied directly to extract deep features in the experiment. 

\subsection{Ablation Study}

\subsubsection{Effect of $\alpha$}
Without the entropy regularization, the obtained distribution will be one-hot vector that is sensitive to small changes. By letting $\alpha=0$, the performance on CIFAR-100 will decrease from $82.9\%$ to $82.1\%$, which confirms the efficacy of the regularization in SeA. 

\subsubsection{Effect of Batch Size}
In our method, the adversarial direction is projected with the examples from the same mini-batch. Although the projection can be decoupled by keeping a memory bank, we observe that the simple setting works well in our experiments. With the parameters obtained for CIFAR-100, the size of the mini-batch is varied in $\{128,256,512,2048\}$, and the performance is summarized in \cref{ta:bs}. All parameters are kept the same except the learning rate, which is scaled according to the batch size (i.e., $\#B$) as $lr = 1\times \#B/256$. 

\begin{table}[h]
\centering
\caption{Comparison of different batch sizes.}\label{ta:bs}
\begin{tabular}{|l|c|c|c|c|}\hline
Batch size&128&256&512&2048\\\hline
Acc (\%)&82.8&82.9&82.9&82.8\\\hline
\end{tabular}
\end{table}

Obviously, the proposed augmentation is robust to different batch sizes. It is because that the examples are randomly sampled at each iteration for SGD, which can capture the whole semantic space well with a sufficient number of examples~\cite{HalkoMT11}. The experiment confirms the effectiveness of the proposed framework with a small batch size.

\subsubsection{Effect of Different Backbones}

\begin{table}[h]
\centering
\caption{Comparison of accuracy ($\%$) with features from ViT pre-trained by MAE~\cite{HeCXLDG22}.}\label{ta:mae}
\resizebox{\linewidth}{!}{
\begin{tabular}{|l|c|c|c|c|c|c|c|c|c|c|c|}\hline
Methods&Aircraft&Birds&Caltech&Cars&Cifar10&Cifar100&DTD&Flowers&Food&Pets&SUN\\\hline
S-R50      &37.4&	66.5&	92.6&	45.8&	89.8&71.3&	72.6&	88.1&	67.8&	92.3&	59.2\\
MAE-ViT& 9.8& 21.0&	81.0&	11.8&	84.3	&57.4&	63.5&	62.3&	52.8&	70.0&	35.0\\\hline
\end{tabular}}
\end{table}
  
Finally, the performance of features from ViT-Base pre-trained by MAE\footnote{\url{https://github.com/facebookresearch/mae}}~\cite{HeCXLDG22} is shown in \cref{ta:mae}. Evidently, compared with supervised pre-trained ResNet-50, semantic information in features extracted from ViT is not well-organized and is ineffective for classification directly, which is consistent with the observation in \cite{HeCXLDG22}. It is because that MAE pre-trains models with a patch-level reconstruction task, which does not capture image-level semantic features. On the contrary, methods in our experiments optimize the image-level learning objective, which is feasible to extract appropriate deep features for downstream tasks.

\end{document}